\documentclass{article}

\usepackage{arxiv}
\usepackage{graphicx}
\usepackage{subcaption}
\usepackage{float}
\usepackage{amsmath, amssymb}

\usepackage{mathtools}
\usepackage{amsthm}
\usepackage{subcaption}

\usepackage[utf8]{inputenc} 
\usepackage[T1]{fontenc}    
\usepackage{hyperref}       
\usepackage{url}            
\usepackage{booktabs}       
\usepackage{amsfonts}       
\usepackage{nicefrac}       
\usepackage{microtype}      
\usepackage{lipsum}		
\usepackage{graphicx}
\usepackage[round]{natbib}
\usepackage{doi}

\title{Optimizing Algorithms for Mobile Health Interventions with Active Querying Optimization}

\author{ Aseel Rawashdeh \thanks{Department of Statistics, Harvard University, Cambridge,  MA 02138. \texttt{arwashdeh@college.harvard.edu}} \\
}

\date{}

\renewcommand{\headeright}{}
\renewcommand{\undertitle}{}

\hypersetup{
pdftitle={Optimizing Algorithms for Mobile Health Interventions with Active Querying Optimization},
pdfsubject={q-bio.NC, q-bio.QM},
pdfauthor={Aseel Rawashdeh},
pdfkeywords={First keyword, Second keyword, More},
}

\begin{document}
\maketitle

\begin{abstract}
Reinforcement learning in mobile health (mHealth) interventions requires balancing intervention efficacy with user burden, particularly when state measurements (for example, user surveys or feedback) are costly yet essential. The Act-Then-Measure (ATM) heuristic addresses this challenge by decoupling control and measurement actions within the Action-Contingent Noiselessly Observable Markov Decision Process (ACNO-MDP) framework. However, the standard ATM algorithm relies on a temporal difference-inspired Q-learning method, which is prone to instability in sparse and noisy environments. In this work, we propose a Bayesian extension to ATM that replaces standard Q-learning with a Kalman filter-style Bayesian update, maintaining uncertainty-aware estimates of Q-values and enabling more stable and sample-efficient learning. We evaluate our method in both toy environments and clinically motivated testbeds. In small, tabular environments, Bayesian ATM achieves comparable or improved scalarized returns with substantially lower variance and more stable policy behavior. In contrast, in larger and more complex mHealth settings, both the standard and Bayesian ATM variants perform poorly, suggesting a mismatch between ATM’s modeling assumptions and the structural challenges of real-world mHealth domains. These findings highlight the value of uncertainty-aware methods in low-data settings while underscoring the need for new RL algorithms that explicitly model causal structure, continuous states, and delayed feedback under observation cost constraints.
\end{abstract}

\section{Introduction}
Mobile health (mHealth) interventions increasingly rely on reinforcement learning (RL) to personalize behavioral treatments over time. A prominent class of applications involves context-aware, just-in-time adaptive interventions (JITAIs) designed to promote physical activity through mobile notifications. In these settings, an RL agent typically faces a dual decision-making problem: (1) whether to deliver a control intervention such as a motivational activity prompt, and (2) whether to measure the user’s internal state (i.e., their momentary attitude toward physical activity) via costly surveys or self-report queries. This introduces a fundamental tradeoff between collecting valuable state information and minimizing user burden, making the design of measurement-aware RL algorithms essential for effective, scalable mHealth systems.

\subsection{Problem Setting and Framework}
From the existing literature, this dual-action setup is most similar to the Action-Contingent Noiselessly Observable Markov Decision Process (ACNO-MDP) framework \citet{nam2021reinforcement}, which is a special class of partially observable Markov decision processes (POMDPs). In ACNO-MDPs, the environment is partially observable unless the agent chooses to take an explicit measurement action, which provides noiseless (i.e., fully accurate) but costly state information. Measurement actions, via sending a survey, incur a cost, namely, user burden, and thus must be limited in frequency. This introduces a fundamental trade-off between state observation and user burden in mHealth interventions.

Traditional methods for solving ACNO-MDPs typically rely on full POMDP solvers, which maintain posterior belief distributions over states and perform value iteration or policy search. However, these approaches are computationally expensive and require large amounts of data due to their reliance on deep RL algorithms, making them impractical in resource- and data-constrained settings such as mobile health, where only small datasets are available and rewards are sparse or delayed.

To address these limitations, \citet{krale2023act} proposed the Act-Then-Measure (ATM) heuristic, which decouples the control and measurement decisions by first selecting the best control action based on the current belief and then deciding whether a measurement is warranted. The ATM framework simplifies policy optimization while maintaining high-quality decision-making in practice.

In this work, we modify the ATM framework by substituting its replicated Q-learning procedure (where each $Q(s,a)$ receives a temporal difference update weighted by the current belief $b(s)$, effectively ‘replicating’ the update across all latent states) with a Bayesian Kalman-filter-style Q-learning rule that tracks uncertainty over Q-values \citep{tripp2013approximate}. This research is motivated by a few important observations:
\begin{itemize}
\item Standard Q-learning suffers from noisy, unstable updates in sparse-reward environments.
\item Kalman-based Q-learning introduces state-action specific uncertainty estimates, allowing more robust learning from limited data.
\item Prior work in continuous-state MDPs and robotic control has shown that Bayesian value estimation improves stability and sample efficiency \citep{engel2005reinforcement, wang2023bayesian}
\end{itemize}

We integrate Bayesian Q-learning into the ATM algorithm and evaluate it across both synthetic and real-world-inspired settings. In particular, we test our approach on the FrozenLake toy environment and a ADAPTS clinical trial testbed, a clinically grounded simulation platform derived from real mHealth trial data. Our results show that while the Bayesian ATM variant achieves improved stability and measurement strategies in small tabular environments, both the standard and Bayesian variants struggle to generalize effectively in the high-dimensional, delayed-reward structure of ADAPTS. his gap illustrates a mismatch between the assumptions of ACNO-MDP algorithms and the structural complexities of real mHealth environments

\subsection{Contributions}
Our contributions are both methodological and empirical:

\begin{itemize}
\item We implement the Act–Then–Measure (ATM) framework in both tabular ACNO-MDP settings and the ADAPTS mHealth trial testbed environment, providing the first end-to-end evaluation of ATM in a realistic clinical trial testbed.
\item We develop a Bayesian extension of ATM by integrating Kalman-style Q-learning, enabling uncertainty-aware value estimation within the ATM control–measurement decomposition.
\item We empirically characterize the benefits and failure modes of Bayesian ATM. In small tabular environments, Kalman Q-learning substantially improves stability and sample efficiency. In contrast, in the ADAPTS testbed, we show that structural features (e.g., partial reward observability, delayed burden, and stochastic engagement dynamics) prevent uncertainty from collapsing, leading Bayesian ATM to over-measure and perform poorly.
\item Through these experiments, we identify a structural mismatch between ACNO-MDP assumptions and real-world mHealth environments, highlighting directions for designing measurement-aware RL algorithms with better causal and statistical grounding.
\end{itemize}

\section{Related work}
\subsection{Active Learning and Cost-Aware Querying in Reinforcement Learning}

Active learning methods aim to reduce the sample complexity of learning by choosing the best actions or querying the most informative data points \citep{settles2009active}. In reinforcement learning (RL), this paradigm has been extended to settings where observations (particularly of rewards or states) are costly, giving rise to active RL frameworks that explicitly model a trade-off between information gain and query cost.

One central line of work formalizes \textit{active reward learning} as a variant of standard RL where the agent must pay a cost to observe rewards. \citet{krueger2020active} introduce the Active Reinforcement Learning (ARL) framework, showing that classical exploration methods like optimism or Thompson sampling can fail under reward observability constraints. Their Mind-Changing Cost Heuristic (MCCH) algorithm uses reward gap estimates to decide whether querying is worthwhile, offering a principled alternative to cost-agnostic querying.

\citet{parisi2024monitored} extend this direction with the Monitored MDP (Mon-MDP) framework, where rewards are only observable through an auxiliary monitor process. Their follow-up work, \citet{parisi2024beyond}, proposes a dual-value architecture to disentangle reward and state uncertainty, emphasizing structured exploration in partially observable reward environments. This work is similar to the belief-based approach in ATM, where rewards (i.e., user health outcomes) are sparse, and measurements must be carefully budgeted.

In another related category, \textit{preference-based active learning} methods elicit feedback by querying trajectory comparisons instead of scalar rewards. \citet{sadigh2017active} propose a method that synthesizes preference queries to maximize expected information gain, effectively narrowing the hypothesis space over possible reward functions. \citet{biyik2020active} extend this work by using Gaussian Process regression to capture non-linear reward functions and by designing acquisition functions that enable batch querying. While these methods assume full observability and differ in query modality, they share a related goal of learning efficiently from limited, costly feedback.

\citet{kong2022provably} offer a theoretically grounded algorithm that separates reward-free exploration from query-based exploitation, querying only when confidence intervals suggest high uncertainty. Unlike this two-phase approach, our method operates under persistent partial observability and integrates querying decisions into every step of learning.

Bayesian techniques are often used to guide active learning via posterior uncertainty. For example, \citet{schulze2018active} adapts Monte Carlo Tree Search to ARL, simulating the long-term benefit of querying. These simulation-based methods highlight the inadequacy of short-sighted exploration bonuses and motivate our use of Kalman-style updates for potentially more robust value estimation.

Recent work also considers \textit{state observability} as a decision variable. \citet{bellinger2022balancing} and \citet{baja2025measure} study settings in which agents selectively measure environmental variables (i.e., robot limb angles or crop growth metrics). These frameworks explicitly trade off measurement costs against performance, as we do, though they often rely on deterministic transitions or known dynamics.

Finally, works like \citet{helmert2022activeobs} and \citet{lqg2021infinite} address continuous control settings in which measurements incur cost, and agents must optimally schedule when to observe. While their control-theoretic framing differs, they reinforce the importance of jointly modeling value estimation and observation cost, as we do in our discrete-state toy environment.

\subsection{Active Query Methods in Reinforcement Learning}

Beyond general active learning approaches, a closely related category of work concerns active querying in reinforcement learning, where the agent must explicitly decide, as part of its policy, whether and when to measure latent variables or observations.

\begin{figure}[t]
    \centering

    \begin{subfigure}[t]{0.48\textwidth}
        \centering
        \includegraphics[width=\linewidth]{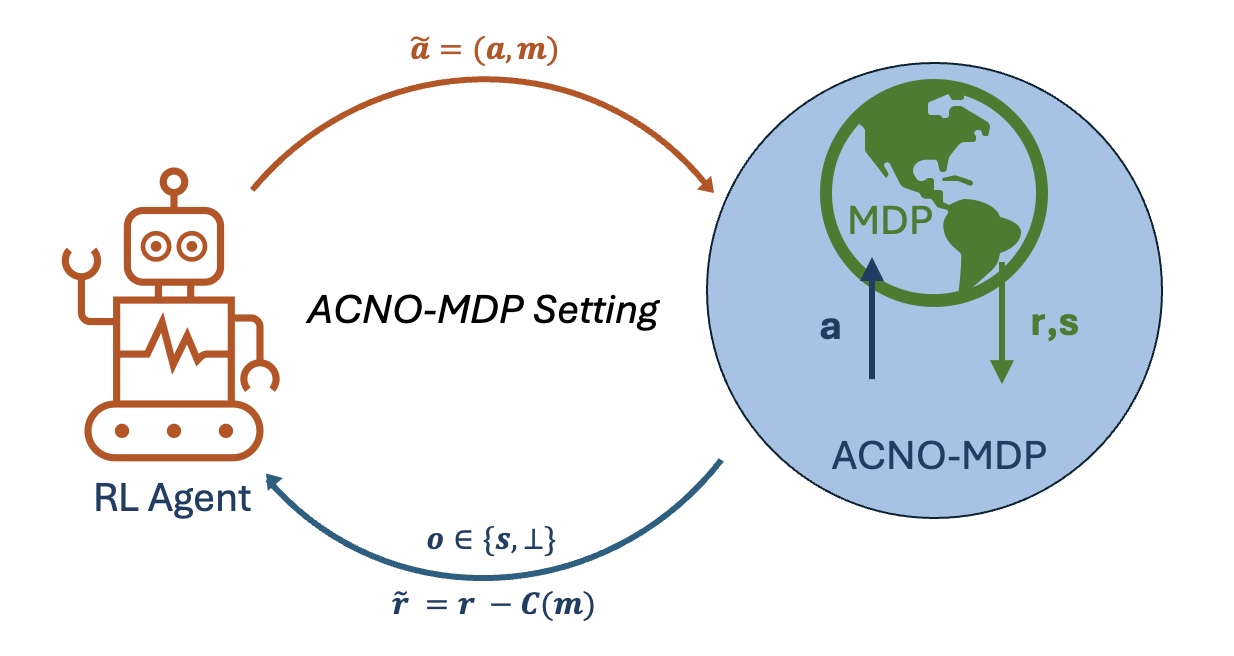}
        \caption{Agent-environment interaction in an ACNO-MDP. The internal environment state is defined by an MDP and affected only by control actions.}
        \label{fig:acno}
    \end{subfigure}
    \hspace{0.4cm}
    \begin{subfigure}[t]{0.48\textwidth}
        \centering
        \includegraphics[width=\linewidth]{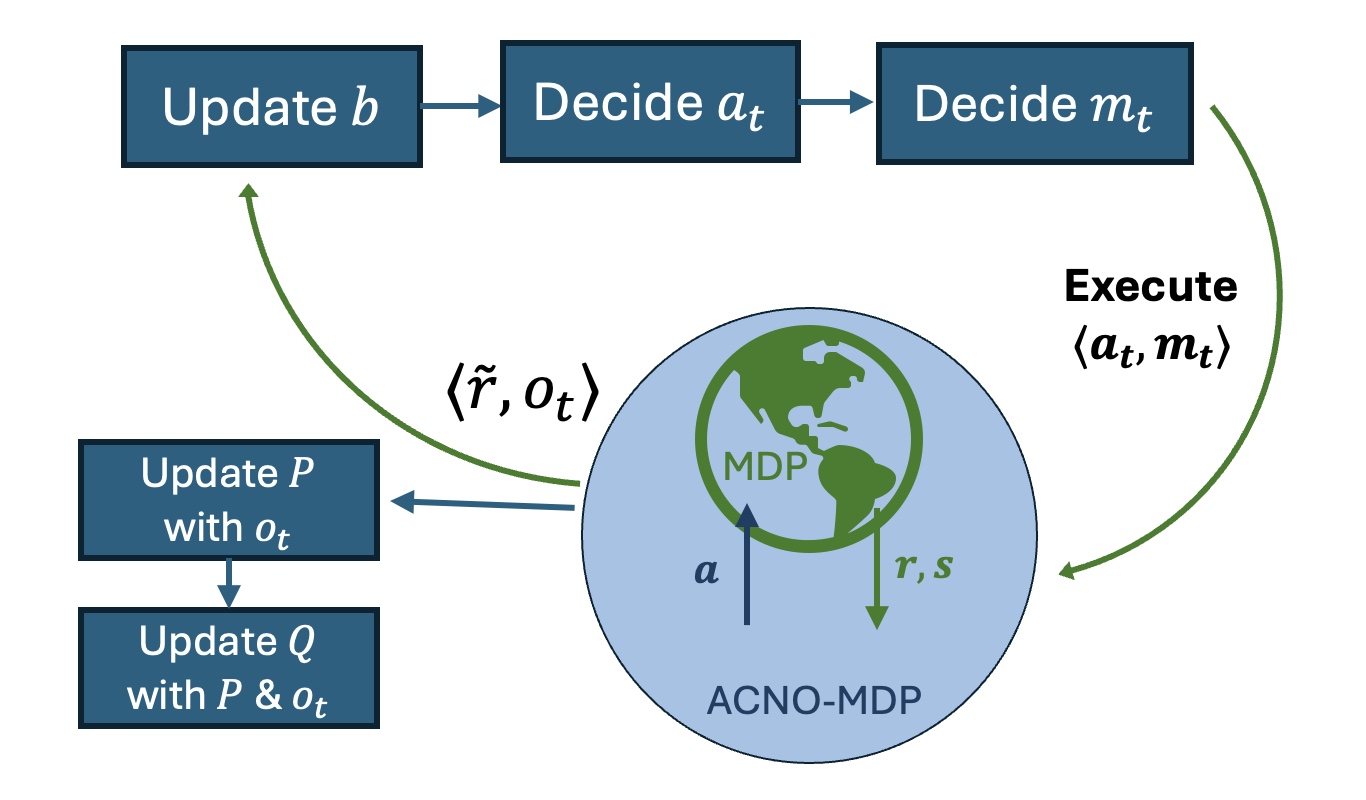}
        \caption{Visual representation of the Act Then Measure Heuristic within the ACNO-MDP framework}
        \label{fig:atm}
    \end{subfigure}
\caption{Graphics illustrating the ACNO-MDP problem setting and ATM heuristic (Figures by Student)}
    \label{fig:framework}
\end{figure}

A foundational framework in this domain is the \textit{Action-Contingent Noiselessly Observable MDP} (ACNO-MDP) setting introduced by Nam et al. \citep{nam2021reinforcement}. In ACNO-MDPs, agents choose both a control action and a binary measurement decision at each step (Fig.~\ref{fig:acno}). The measurement decision incurs a cost but yields a complete, noiseless observation of the environment state. Nam et al. proposed two main algorithms for this setting: \textit{Observe-then-Plan}, which performs full measurement during an initial phase to build a model, then applies a POMDP planner; and \textit{Observe-while-Planning}, which interleaves model building and planning online. While both achieve strong performance in small environments, their reliance on computationally expensive POMDP solvers limits their scalability and adaptability. Moreover, Observe-then-Plan requires a costly up-front measurement phase, making it less suitable for dynamically changing settings, in both mHealth and beyond.

Another method, \textit{AMRL-Q} (Active Measure Reinforcement Learning Q-learning), proposed by Bellinger et al., explicitly incorporates measurement costs into a tabular Q-learning framework \citep{bellinger2021active}. At each step, the agent selects both an environment action and whether to pay to observe the next state. Over time, AMRL-Q uses a learned transition model to replace costly measurements with model-based estimates, thereby improving cost efficiency. However, AMRL-Q suffers from two important limitations: (1) it consistently converges to non-measuring policies regardless of whether measuring is optimal, and (2) it estimates future value using only the most-likely next state, ignoring belief uncertainty. These limitations degrade performance in stochastic settings where measurements must be taken periodically to re-anchor the agent’s belief state: without intermittent ground-truth observations, uncertainty compounds over time, the learned transition model becomes inaccurate, and value estimates diverge from true dynamics.

In contrast, the \textbf{Act-Then-Measure} heuristic proposed by Krale et al. addresses the shortcomings of AMRL-Q and Observe-then-Plan by decoupling action and measurement selection \citep{krale2023act}. ATM first selects the best control action under the assumption of full observability ($Q_{MDP}$-style, inspired by \citet{littman1995learning}), then uses a belief-based metric (referred to as the measuring value, MV) to determine whether a measurement is warranted (Fig.~\ref{fig:atm}). This design enables ATM to adaptively gather information when needed without incurring excess or unnecessary costs. The authors implement this approach as Dyna-ATMQ, a model-based Q-learning algorithm with belief updates and optional offline training. Empirically, ATM outperforms AMRL-Q and Observe-then-Plan in both synthetic and benchmark ACNO-MDP environments, achieving higher scalarized return and better scalability, especially in smaller environments.

Despite its advantages, ATM has limitations that constrain its application in specific settings. Most notably, it assumes that uncertainty will be resolved after the next transition, an assumption that can lead to inaccurate value estimates and suboptimal behavior in delayed or partially observable environments. Additionally, the measurement decision is decoupled from the control action selection process, meaning long-term measurement consequences are not considered during control optimization. Like earlier ACNO-MDP approaches, ATM also struggles to extend to continuous or high-dimensional state spaces and does not directly handle scenarios where query actions affect the environment dynamics or incur delayed costs—both common features in certain mHealth interventions.

Our work builds directly onto the ATM method by incorporating \textit{Bayesian Q-learning with Kalman-style updates} into the ATM framework. Thus, we position our contribution within the active query literature: we refine the ATM heuristic with principled uncertainty-aware updates while critically examining its assumptions and extending its applicability to domains where uncertainty, environment dynamics, and cost are deeply intertwined.

\section{Methodology}

This section outlines the methods used in our work. We begin by reviewing the Act-Then-Measure (ATM) heuristic and its original replicated Q-learning formulation for solving ACNO-MDPs. We then introduce our Bayesian Q-learning extension, motivated by Kalman filtering, and describe its theoretical formulation and practical benefits.

\subsection{Standard Act-Then-Measure with Replicated Q-Learning}

The ATM heuristic, originally introduced by \citet{krale2023act}, is designed to simplify decision-making in ACNO-MDPs by separating the control and measurement decisions. At each time step, the agent first selects a control action $a$ based on its current belief $b$ over states (a distribution over the states representing the probability of being in each state of the environment), and then decides whether to take a costly measurement ($m \in \mathcal{M} = \{\text{observe}, \text{not observe}\}$). This separation is based on the assumption that the uncertainty in the belief state will be resolved after the measurement, allowing a simplification of the Q-value computation. After executing the action-pair $(a, m)$ in state $s$, the environment transitions to a new state and returns to the agent a reward $r$, a cost $c = C(m)$, and possibly an observation. The goal of the agent is to compute a policy $\pi$ with the highest expected discounted scalarized reward $V(\pi, \mathcal{M}) = \mathbb{E}_{\pi,\mathcal{M}}[\sum_t \gamma^t \tilde{r}_t]$, where $\tilde{r}_t = r_t - C(m_t)$. It is assumed that the agent only has access to the number of states and returned signals, but otherwise has no prior knowledge of the environment dynamics.

\paragraph{Belief-state transitions.}  
When the agent does not measure, it updates its belief using the predictive rule  
\[
b'(s' \mid b,a) = \sum_{s \in \mathcal{S}} b(s) P(s' \mid s,a).
\]
If a measurement is taken, the belief collapses to a one-hot distribution over the observed state. This belief update mechanism is central to evaluating both action values and the usefulness of measurements.

\paragraph{Evaluating measurement decisions.}  
ATM determines whether a measurement is beneficial by comparing the Q-value of measuring with the Q-value of not measuring:
\[
MV(b,a) = Q_{\text{ATM}}(b, \langle a,1\rangle) - Q_{\text{ATM}}(b,\langle a,0\rangle),
\]
where measuring allows the agent to condition its next action on the revealed true state. A key intermediate quantity is the belief-optimal control action
\[
\tilde{a}_b = \arg\max_{\tilde{a} \in \mathcal{A}} Q_{\text{ATM}}(b_{\text{next}}(b,a), \tilde{a}),
\]
which reflects the optimal control choice after either observing or predicting the next belief. The ATM measuring rule then takes the form
\[
m = 1 \;\; \text{if and only if} \;\; MV(b,a) \ge 0.
\]

\paragraph{Control-value approximation.}  
Because computing $Q_{\text{ATM}}$ exactly for all belief states is computationally prohibitive, the ATM method approximates the control Q-value as
\[
Q(b, a) = \sum_{s \in \mathcal{S}} b(s) \, Q_{\text{MDP}}(s,a),
\]
where $Q_{\text{MDP}}(s,a)$ denotes the optimal MDP Q-value under known dynamics. This linear approximation is justified by the ATM assumption that control and measurement decisions can be separated.

\paragraph{Replicated Q-learning.}  
To learn in unknown environments, ATM uses replicated Q-learning, which maintains and updates a separate Q-value $Q(s,a)$ for each underlying state. Transition estimates are updated only when the agent measures:

$$\alpha_{s,a,s'} \leftarrow \alpha_{s,a,s'} + \mathbf{1}\{\text{state } s' \text{ observed}\},$$ 

$$P(s'|s,a) = \frac{\alpha_{s,a,s'}}{\sum_{s''} \alpha_{s,a,s''}}.$$

Given these transition estimates, replicated Q-learning updates $Q(s,a)$ using a belief-weighted temporal-difference rule:
\[
Q(s, a) \leftarrow (1 - \eta_s) Q(s, a) + \eta_s \left(\tilde{r} + \gamma \Psi(s, a) \right),
\]
where
\[
\eta_s = b(s)\eta, \qquad 
\Psi(s,a) = \sum_{s'} P(s'|s,a) \max_{a'} Q(s',a').
\]
This method assumes a point estimate for each $Q(s,a)$ and does not explicitly track uncertainty.

\paragraph{Exploratory and exploitative measurements.}  
ATM further distinguishes between two classes of measurements. Early in learning, the agent performs \emph{exploratory} measurements for the first $N_m$ visits to each state–action pair to ensure accurate transition estimates. Once basic transition knowledge is acquired, the agent switches to \emph{exploitative} measurements driven by the measuring value $MV(b,a)$, which determine whether resolving uncertainty improves expected return.

\paragraph{Performance guarantees.}  
Under mild conditions on dynamics and measurement cost, the ATM heuristic enjoys a bounded regret relative to the optimal joint policy. This guarantee relies on the fact that, given the approximated control rule, the measurement rule $MV(b,a)\ge 0$ is locally optimal. These guarantees motivate ATM as a principled starting point; in the Bayesian extension that follows, this point estimate is replaced with a posterior mean and variance.

\subsection{Bayesian Kalman Q-Learning for ATM}

To address the instability and overconfidence that arise in small-sample or noisy environments, we replace the standard replicated Q-learning update with a Bayesian Q-learning formulation using Kalman filter principles \citep{tripp2013approximate}. In this approach, each $Q(s,a)$ maintains both a posterior mean and variance, enabling uncertainty-aware value estimation.

We model each $Q(s,a)$ as a Gaussian random variable:
\begin{equation}
Q(s,a) \sim \mathcal{N}(\mu_{s,a}, \sigma^2_{s,a}).
\end{equation}

Upon observing a transition $(s, a, r, s')$, the updates are:
\begin{itemize}
\item \textbf{Prediction (Target):}
\begin{equation}
\nu(s, a) = r + \gamma \max_{a'} \mu_{s',a'}.
\end{equation}

\item \textbf{Kalman Gain:}
\begin{equation}
K = \frac{\sigma^2_{s,a}}{\sigma^2_{s,a} + \tau^2},
\end{equation}

\item \textbf{Mean Update:}
\begin{equation}
\mu_{s,a} \leftarrow \mu_{s,a} + K (\nu(s,a) - \mu_{s,a}),
\end{equation}

\item \textbf{Variance Update:}
\begin{equation}
\sigma^2_{s,a} \leftarrow (1 - K) \sigma^2_{s,a}.
\end{equation}
\end{itemize}
Here, $\tau^2$ is the observation noise variance, typically set as a small positive value (e.g., $0.1$ to $0.5$).

\paragraph{Motivation.}
Kalman-style Q-learning introduces uncertainty-aware updates: uncertain state–action pairs receive larger updates, while confident pairs are adjusted conservatively. This adaptivity improves sample efficiency and stability, especially in sparse-reward domains like mHealth, where agents make infrequent measurements, and transitions are often only partially observed.

\paragraph{Integration with ATM.}
To integrate Bayesian Q-learning into the ATM framework, we replace all point-valued $Q(s,a)$ estimates with their posterior means $\mu_{s,a}$ in both the control-value approximation and the measurement-value computation. The belief-weighted control Q-value thus becomes:
\begin{equation}
Q(b,a) = \sum_{s} b(s)\,\mu_{s,a},
\end{equation}
mirroring the linear approximation used in standard ATM.

For measurement decisions, the Bayesian version evaluates
\[
MV(b,a) = Q_{\text{ATM}}^{\text{Bayes}}(b,\langle a,1\rangle) - Q_{\text{ATM}}^{\text{Bayes}}(b,\langle a,0\rangle),
\]
where each term depends on the posterior means and transition estimates. Because the posterior variance $\sigma^2_{s,a}$ captures epistemic uncertainty, the agent implicitly accounts for belief drift and model error when determining whether information is worth acquiring.

We retain the ATM assumption that control decisions are made as if uncertainty will be resolved in the next timestep, but uncertainty now affects value estimates and the measuring threshold through the Kalman gain and posterior variance.

\paragraph{Predicted Advantages.}
\begin{itemize}
\item Improved stability in sparse-reward or noisy regimes
\item More robust value estimation under limited measurements
\item Natural principled exploration through posterior variance
\end{itemize}

This Bayesian extension aligns with methods in \citet{che2021bayesian} and \citet{tripp2013approximate}, but is adapted for belief-based decision-making in ACNO-MDPs, including explicit modeling of measurement cost, belief updates, and uncertainty-aware replicated Q-learning.

\section{Experiments}

To evaluate the efficacy of Bayesian Q-learning within the Act-Then-Measure framework, we begin with a similar experimental setup to that proposed by \citet{krale2023act}, using similar environments but replacing original comparisons with a direct comparison between two ATM-based variants:

\begin{itemize}
\item \textbf{ATM-Q (Replicated Q-learning):} Implements ATM with the standard (TD-inspired) replicated Q-learning, maintaining a single Q-value estimate per state-action pair.
\item \textbf{ATM-KQ (Kalman Q-learning):} A Bayesian extension of ATM-Q that maintains a Gaussian posterior for each Q-value, updates parameters via Kalman-style rules, and uses posterior means for both control selection and measurement value computation.
\end{itemize}

Each agent is evaluated over 5 independent runs with a set number of episodes (defined for each experiment).

\subsection{Environments}

First, we test both agents in a few canonical ACNO-MDP environments:

\paragraph{Measuring Value Environment.} This is a minimal tabular environment composed of three states: $s_0$, $s_+$, and $s_-$, where the agent starts in $s_0$ and transitions to $s_+$ or $s_-$ probabilistically upon taking action $a_1$. Rewards $r=1$ or $r=0$ are returned, respectively, but the agent must decide whether to pay a cost $c$ to observe which of the indistinguishable states $s_+$ or $s_-$ it has entered. We vary the measurement cost $c$ from $0.04$ to $0.20$ and evaluated the average performance over $1000$ episodes.

\paragraph{FrozenLake Environment.} The \texttt{OpenAI Gym} FrozenLake environment is a grid-based navigation task in which an agent must learn to reach a goal state from a fixed starting position while avoiding hazardous “holes” that end the episode prematurely. The agent receives a reward of $r=1$ upon successfully reaching the goal, and $r=0$ otherwise. At each time step, it chooses an action from the set $\{\texttt{left}, \texttt{right}, \texttt{up}, \texttt{down}\}$ and navigates the environment based on an underlying transition model that may include stochasticity. To adapt this environment to the ACNO-MDP setting required by ATM, we make the agent's location partially observable and allow it to query its true position at cost $c$. This modification is necessary because the standard FrozenLake is fully observable, but ATM requires partial observability with optional measurement. The agent maintains a belief over grid cells and updates the belief using the predictive rule introduced in the Methodology section.

We evaluate on both standard and custom variants of FrozenLake with a fixed measurement cost $c=0.05$, extending the environment to support optional querying of the current state. In all cases, ATM-Q and ATM-KQ update transition estimates and correct belief states only when a measurement is taken. We use $4\times4$ maps for our small-scale experiments and randomly generated $n\times n$ maps for larger-scale experiments, with $n \in [8, 20]$. For each setting, the agent is evaluated for $1000$ episodes (small lakes) or $7500$ episodes (large lakes).

To analyze behavior under different transition dynamics, we consider three settings: (1) \textbf{deterministic}, where the agent moves precisely in the intended direction; (2) \textbf{slippery}, where movement is randomized with equal probability of deviating perpendicularly; and (3) \textbf{semi-slippery}, a middle ground in which the agent moves in the desired direction, but with a $0.5$ probability of moving two steps instead of one. These variants allow us to simulate increasing levels of uncertainty, mirroring the transition unpredictability and noisy feedback often observed in real-world mHealth environments.

Second, we test the agents in the ADAPTS trial testbed: 
\begin{figure}[t]
\centering

\includegraphics[width=\linewidth]{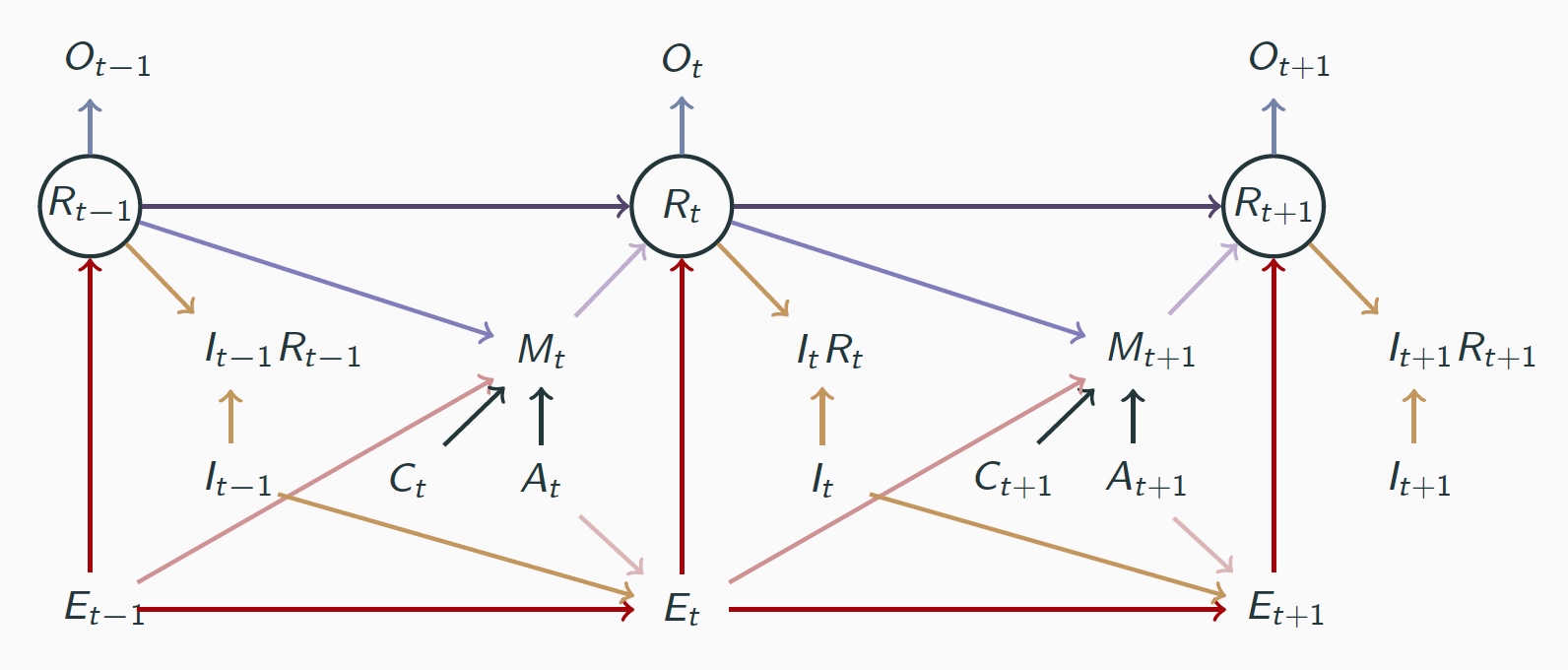}
\caption{Bag length $K = 1$. Arrows pointing to the actions are omitted. $A_t$: control action, $I_t$: query action, $R_t$: latent reward, $I_t R_t$: revealed reward, $C_t$: context, $M_t$: mediator, $E_t$: engagement, $O_{t-1}$: observation of $R_t$}
    \label{fig:dag}
\end{figure}

\paragraph{ADAPTS Testbed.}
The ADAPTS testbed is derived from HeartSteps V2, a clinical trial designed to promote long-term physical activity through context-aware mobile interventions. Each simulated user is parameterized using individualized estimates from real HeartSteps data, enabling realistic modeling of behavior under different RL policies. The environment is structured into daily “bags’’ of $K=5$ decision points, where the agent selects whether to deliver a physical activity suggestion ($A_t$). The mediator $M_t$ (recent step count) captures the short-term behavioral effect of the intervention, while the latent reward $R_t$ reflects user attitude toward physical activity and is revealed only when a survey query is sent. Context $C_t$ and engagement $E_t$ are always observable, but $R_t$ is only observed when measured, making the measurement action the mechanism through which the agent recovers the unobserved portion of the state.

To apply ATM and its Bayesian extension in this setting, we constructed a tabular abstraction using a 64-state discretization of $(C_t, E_t, R_t)$ based on standardized normal-based binning; the discretization is required because the underlying ATM algorithm is defined for tabular state spaces and cannot directly operate on the continuous features of the ADAPTS environment. This representation preserves the mixed continuous–discrete structure of the environment while enabling belief updates, Dirichlet transition estimation, and Q-learning within the ATM framework. We also parameterized the measurement cost, discount factor, and offline training steps to match the ADAPTS time horizon and survey burden. A terminal “done’’ state was added to handle episode termination, and we separated warm-up and intervention phases to reflect how real-world mHealth systems collect initialization data before deploying a policy. These adaptations allow the core ATM decision rule (belief-weighted control updates combined with a separate measurement decision) to be embedded cleanly within a complex mHealth environment.

\paragraph{Causal Dynamics and Reinforcement Learning Challenges.}
The ADAPTS testbed presents several causal and statistical complexities that make it substantially more challenging than standard ACNO-MDP benchmarks. First, query actions (i.e., sending surveys) directly influence the environment by altering engagement $E_t$, which mediates future rewards $R_t$; thus, querying introduces delayed and stochastic burden rather than an immediate scalar cost. This violates the simplifying assumption used by many active-measurement algorithms, including ATM, in which the measurement cost does not modify the underlying transition dynamics. Second, because the reward $R_t$ is latent and observed only under measurement, the agent must simultaneously learn a transition model, infer unobserved state components, and reason about the long-term consequences of querying. Third, while context and engagement are always observable, the reward $R_t$ is latent and missing unless queried, complicating credit assignment and belief propagation.

The high-dimensional, continuous nature of the original testbed necessitates using a coarse 64-state discretization of $(C,E,R)$. While this abstraction enables belief updates and tabular Q-learning, it inevitably loses behavioral nuance and creates sparse state–action visitation, amplifying the data scarcity challenges inherent to mHealth. As our results show, these factors collectively hinder the performance of both standard ATM and its Bayesian extension in ADAPTS, highlighting the difficulty of applying active-measurement RL algorithms in environments with causal feedback loops, delayed burden, and partially observed latent preferences. We emphasize that the modeling choices discussed in this section reflect limitations of the ATM framework rather than inherent properties of the ADAPTS testbed, which is originally formulated in continuous state space.

\subsection{Evaluation Metrics}

For each experiment, we track:
\begin{itemize}
\item \textbf{Scalarized return:} Reward minus measurement cost over episodes ($\tilde{r} = r- C(m)$)
\item \textbf{Measurement frequency:} Average number of measurements taken
\item \textbf{Step count:} Average number of environment steps until episode termination
\end{itemize}

Results are averaged across 5 runs, and plots include 95\% confidence intervals. Additional visualizations (i.e., scalarized return vs. measurement cost) are reported in Appendix~\ref{app:mc_plots}.

These experiments allow us to assess the sample efficiency, stability, and query behavior of Kalman-style Bayesian Q-learning within a complex toy environment.

\section{Results}

\begin{figure}[t]
\centering
\begin{tabular}{lccc}
\toprule
\textbf{Algorithm} & \textbf{Cost} & \textbf{SR} & \textbf{M} \\
\midrule
ATM-Q & 0.05 & 0.53 & 5.65 \\
ATM-Q + Kalman QL & 0.05 & \textbf{0.64} & \textbf{6.42} \\
\bottomrule
\end{tabular}
\caption{Average scalarized return (SR) and number of measurements (M) in the measuring value environment ($c=0.05$), averaged over the final 200 episodes across 5 runs.}
\label{fig:mv_table}
\end{figure}

\begin{figure}[t]
\centering
\resizebox{\columnwidth}{!}{%
\begin{tabular}{lcccccc}
\toprule
\textbf{Algorithm} & \textbf{SR (Det)} & \textbf{SR (Semi)} & \textbf{SR (Slip)} & \textbf{M (Det)} & \textbf{M (Semi)} & \textbf{M (Slip)} \\
\midrule
ATM-Q & 1.00 & 0.53 & \textbf{-0.17} & 0.00 & 5.65 & 4.14 \\
ATM-Q + Kalman QL & 1.00 & \textbf{0.64} & -3.64 & \textbf{0.04} & \textbf{6.42} & \textbf{83.87} \\
\bottomrule
\end{tabular}
}
\caption{Scalarized return (SR) and number of measurements (M) for ATM-Q variants in FrozenLake variants ($c=0.05$), averaged over the final 200 episodes.}
\label{fig:lake_table}
\end{figure}

\begin{figure}[t]
    \centering

    \begin{subfigure}[t]{0.48\textwidth}
        \centering
        \includegraphics[width=\linewidth]{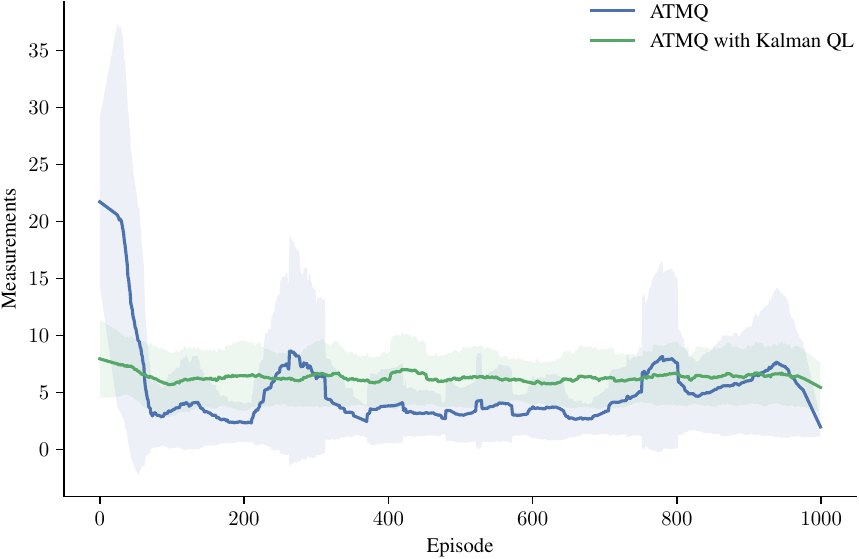}
        \caption{Measurement count over episodes in semi-slippery environment.}
        \label{fig:measures}
    \end{subfigure}
    \hfill
    \begin{subfigure}[t]{0.48\textwidth}
        \centering
        \includegraphics[width=\linewidth]{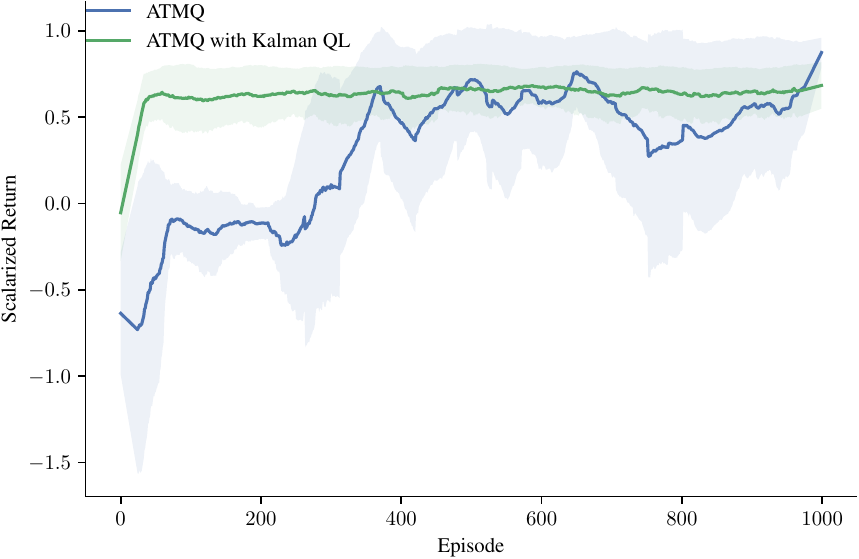}
        \caption{Scalarized return over episodes in semi-slippery environment.}
        \label{fig:sr}
    \end{subfigure}

    \vspace{0.4cm}

    \begin{subfigure}[t]{0.48\textwidth}
        \centering
        \includegraphics[width=\linewidth]{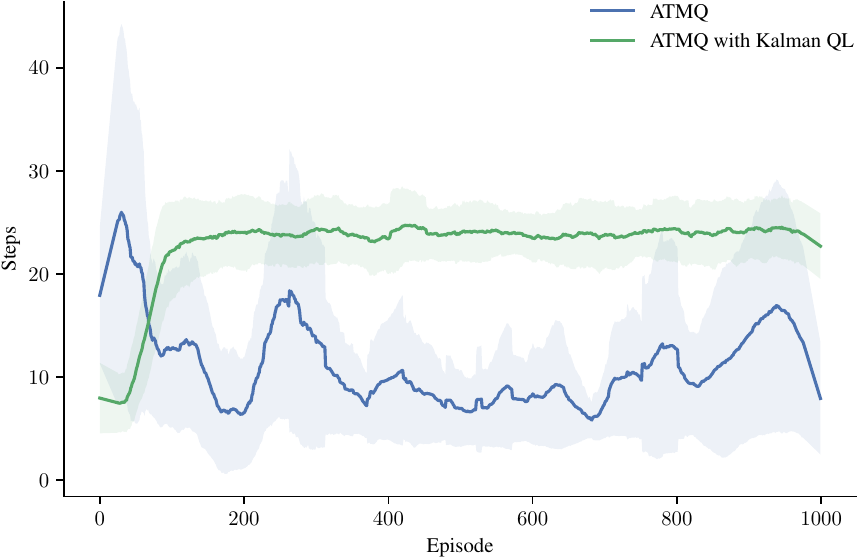}
        \caption{Step count over episodes in semi-slippery environment.}
        \label{fig:steps}
    \end{subfigure}
    \hfill
    \begin{subfigure}[t]{0.48\textwidth}
        \centering
        \includegraphics[width=\linewidth]{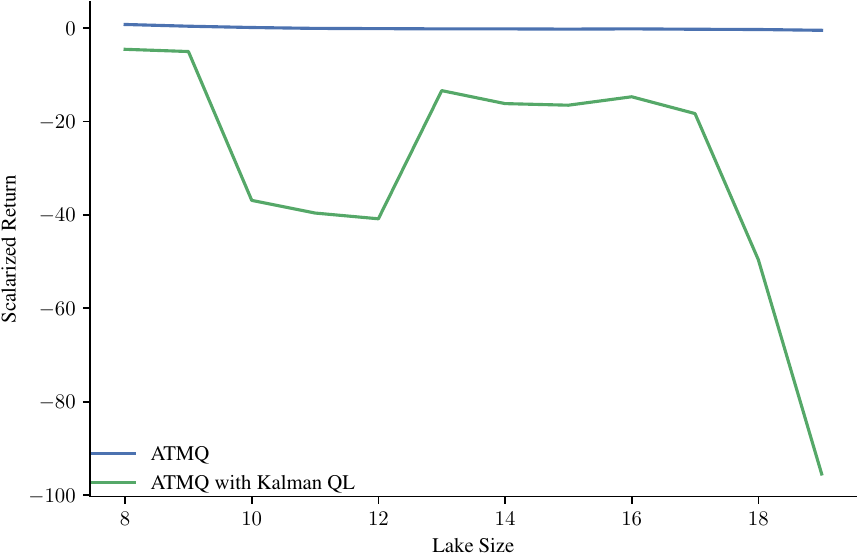}
        \caption{Scalarized return vs lake size in large semi-slippery environments.}
        \label{fig:large_lake}
    \end{subfigure}

    \caption{Performance of ATMQ vs Bayesian ATMQ across different evaluation settings.}
    \label{fig:comparison}
\end{figure}

We evaluate the performance of standard ATM-Q learning and our proposed Bayesian / Kalman style Q-learning variant across multiple environments, focusing on evaluating learning stability, measurement behavior, and scalability under uncertainty.

\subsection{Small Frozen Lake (Semi-Slippery)}

Figure~\ref{fig:sr} shows the scalarized return over 1000 episodes in the semi-slippery 4$\times$4 FrozenLake environment. We observe that the Kalman Q-learning (KQL) variant exhibits several advantages over the standard baseline:

\begin{itemize}
\item \textbf{Lower variance in returns}: The green curve (ATM-KQL) maintains tight confidence bounds throughout training, indicating that the Bayesian updates are less susceptible to stochastic fluctuations. This is especially important in sparse-reward environments, where the noise in value estimates can lead to erratic behavior if not properly regulated.

\item \textbf{Faster convergence and higher stability}: ATM-KQL reaches a stable policy within the first 100 episodes, while the standard ATM-Q baseline continues to oscillate due to unstable TD-style updates. This is because the Kalman-style update mechanism tempers the learning rate based on state-action uncertainty, enabling more stable convergence. 

\item \textbf{Slightly higher average performance}: Averaged over the final 200 episodes,  ATM-KQL achieves a scalarized return of 0.64 compared to 0.53 for ATM-Q (As seen in Fig. ~\ref{fig:mv_table}). While the difference is modest, it is consistent as seen in Figure ~\ref{fig:sr} and reinforces the benefit of incorporating uncertainty-aware updates to navigate partial observability and stochastic feedback more effectively.
\end{itemize}

\subsection{Steps and Measurements}

Figures~\ref{fig:steps} and~\ref{fig:measures} analyze the number of steps and measurements per episode. We observe that:

\begin{itemize}
\item \textbf{Higher and more stable step counts}:  ATM-KQL maintains an average of around 23 steps per episode, with minimal variance. This indicates that the Bayesian agent takes longer but more deliberate paths, reflecting a better understanding of the long-term value of actions. In contrast, ATM-Q shows highly variable step counts and unstable behavior, often prematurely terminating episodes or oscillating between exploratory paths, which may signal overfitting to short-term rewards.

\item \textbf{Slightly higher measurement frequency}:  ATM-KQL consistently performs more measurements (6.42 vs. 5.65 on average) as seen in Figures ~\ref{fig:measures} and ~\ref{fig:mv_table}. This is a direct consequence of its explicit modeling of uncertainty: when posterior variance is high, the agent is more inclined to pay the measurement cost to reduce ambiguity. Notably, this behavior persists even late in training, suggesting that the Bayesian agent selectively queries to refine high-uncertainty regions, rather than blindly exploring.
\end{itemize}

\subsection{Extreme Case: Slippery Environment}

In the fully stochastic (slippery) FrozenLake environment, both methods struggle, but  ATM-KQL performs markedly worse. It obtains a negative scalarized return (-3.64 vs. -0.17) and executes an excessive number of measurements (83.87 vs. 4.14), despite a fixed measurement cost (Fig. ~\ref{fig:lake_table}).

This behavior is explained as follows:

\begin{itemize}
\item \textbf{Measurement overactivation under high uncertainty}: In highly stochastic environments, transition and reward noise leads to high posterior variance in the Q-value estimates.  ATM-KQL, which uses this variance to modulate its Kalman gain, interprets the environment as persistently uncertain and thus overestimates the value of measuring. This results in frequent, expensive queries that do not significantly improve value estimation.

\item \textbf{Lack of variance regularization}: The Bayesian update mechanism does not include a cap on variance growth or decay rate. As such, in low signal-to-noise regimes, the agent’s uncertainty may remain high indefinitely, causing it to mistakenly allocate resources toward repeated measurements even when it is no longer beneficial.

\item \textbf{Detrimental feedback loop}: Frequent measurements without meaningful return lead to a negative scalarized return, which further skews the agent’s posterior estimates. Without a mechanism to recover from such feedback loops (i.e., via posterior smoothing or regret-based querying), the Bayesian version remains trapped in a detrimental suboptimal behavior pattern.
\end{itemize}

\subsection{Large Frozen Lake Performance}

To evaluate scalability, we extend both algorithms to increasingly large 2D FrozenLake grids ranging from $8\times8$ to $20\times20$, using the semi-slippery variant and a measurement cost of $c=0.05$. Figure~\ref{fig:large_lake} presents the scalarized return after convergence.

We observe the following:

\begin{itemize}
\item \textbf{Performance degradation with environment size}: As the state space expands, both algorithms experience a drop in performance, but  ATM-KQL declines more rapidly. Notably, computation times are increased significantly, likely due to the higher computational cost attributed to Kalman-style updates. This highlights a key limitation of the KQL-based ATM implementation in high-dimensional MDPs.

\item \textbf{Exploration bottlenecks due to sample sparsity}: In very large environments, the agent rarely revisits the same $(s,a)$ pairs enough to reduce posterior variance. This causes the Bayesian learner to remain overly cautious and underconfident, inhibiting generalization and leading to inefficient exploration.

\item \textbf{Value overestimation in underexplored regions}: The Bayesian variant may also retain high variance estimates across many unvisited states, falsely identifying them as promising measurement targets. This further drains resources without gaining reward, contributing to a rapid collapse in scalarized return.

\item \textbf{ATM-Q stability advantage}: Despite being less principled, ATM-Q’s simpler point estimates prove more robust in large-scale settings. It quickly commits to approximated Q-values without accumulating variance or increasing computational demands, allowing it to form a rough but workable policy early in training.
\end{itemize}

Taken together, our results show that the Bayesian Q learning variant of ATM offers clear benefits in smaller, noisy, and data-limited environments, but faces critical challenges in scaling to high-dimensional environments without additional support.

\subsection{Performance in the ADAPTS Testbed}

\begin{figure}[t]
    \centering

    \begin{subfigure}[t]{0.48\textwidth}
        \centering
        \includegraphics[width=\linewidth]{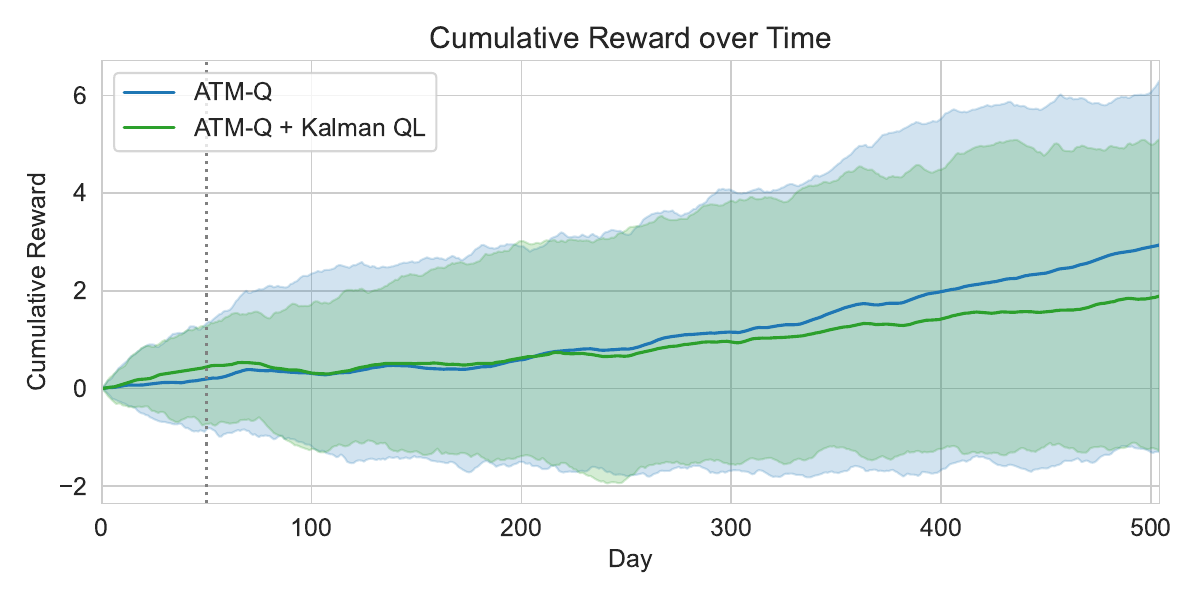}
        \caption{Cumulative reward over time.}
        \label{fig:testbed_reward}
    \end{subfigure}
    \hspace{0.4cm}
    \begin{subfigure}[t]{0.48\textwidth}
        \centering
        \includegraphics[width=\linewidth]{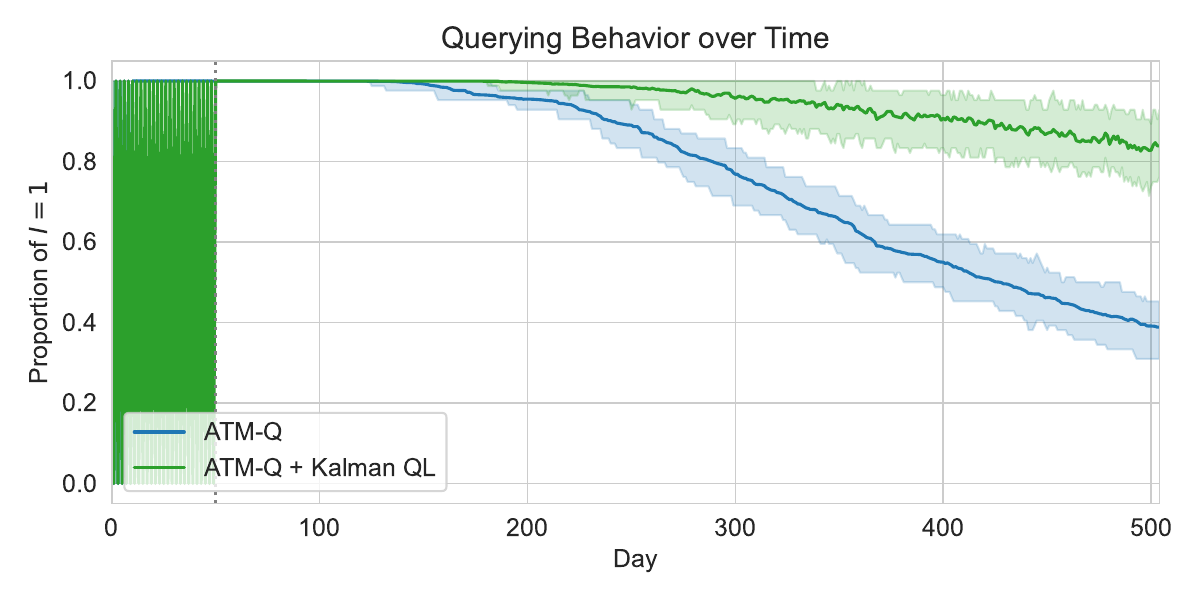}
        \caption{Proportion of queries over time across all users.}
        \label{fig:testbed_query}
    \end{subfigure}
\caption{Results from running the standard ATM method and the modified Kalman-QL method in the ADAPTS testbed}
    \label{fig:adapts_testbed}
\end{figure}

To evaluate the generalizability of ATM and its Bayesian Q-learning extension in realistic mHealth settings, we test both algorithms in the ADAPTS testbed, a simulation environment derived from real-world mobile health intervention data. 

Figure~\ref{fig:testbed_reward} presents the cumulative reward over 500 decision days. Both ATM-Q and ATM with Kalman QL achieve slight positive rewards but exhibit high variance and no sustained upward trajectory in learning. ATM-Q performs marginally better overall, but the large overlap in confidence intervals suggests that this difference is not significant. This plateauing behavior implies that both agents struggle to discover stable, high-reward policies in the face of delayed and sparse feedback.

Figure~\ref{fig:testbed_query} illustrates the average proportion of times the agent chooses to query ($I=1$ or $m=1$). Notably, ATM-Q learns to query less frequently over time, eventually reducing to a stable rate slightly below 0.4. In contrast, ATM with Kalman QL maintains a consistently high query rate (above 0.8) throughout the experiment. This suggests that while the Bayesian variant maintains uncertainty-awareness, it fails to effectively distinguish when querying is beneficial versus redundant, potentially due to overestimated variance in its Q-values or insufficiently informative state representations. This resonates with our findings from the large lake toy environments.

A key driver of this over-measuring behavior is that the posterior variance over $Q(s,a)$ remains high in ADAPTS, even after substantial training. Unlike tabular ACNO-MDP environments, where measurements reveal the full latent state and thus rapidly collapse uncertainty, ADAPTS provides only partial information about the user’s underlying preference dynamics, and this information is confounded by stochastic engagement transitions. As a result, the Bayesian agent continually receives noisy or weakly informative updates, preventing the posterior from concentrating. Since the Kalman gain depends inversely on posterior variance, persistently high uncertainty drives ATM-KQ to assign large informational value to measurements, causing the agent to query at disproportionately high rates. This mechanism clarifies why Bayesian ATM improves stability in controlled tabular environments but performs suboptimally in ADAPTS, where uncertainty does not collapse, and measurements only weakly identify the latent reward.

Together, these results suggest that both algorithms underperform in this testbed. A major limitation is their reliance on a discretized state space, which likely discards subtle structure in the high-dimensional, continuous state characteristics of real-world clinical settings. Also, the assumption that query actions only reveal state (usually valid in tabular MDPs) does not hold in ADAPTS, where querying itself affects downstream engagement and reward. Additional complications include delayed costs, temporally decoupled action-selection, and data scarcity. These properties violate key assumptions of standard ATM formulations, even when modified to incorporate uncertainty.

This highlights the need for new algorithms that can:
\begin{itemize}
\item Jointly model the causal impact of measurements on dynamics and reward
\item Operate on continuous states without relying on hand-crafted discretization
\item Effectively propagate uncertainty over long horizons with minimal data
\end{itemize}
Ultimately, these insights motivate future work on principled query-aware decision-making under partial observability for personalized mHealth interventions.

\section{Discussion and Conclusion}

Our experiments demonstrate that integrating Bayesian Q-learning into the Act-Then-Measure framework introduces meaningful gains in learning stability and robustness under uncertainty, particularly in tabular sparse-reward, partially observable environments.

In small, semi-stochastic domains like the $4 \times 4$ semi-slippery FrozenLake environment, our Bayesian variant learns more stable policies and exhibits significantly lower variance in scalarized return. These improvements stem directly from the Kalman-style updates, which temper noisy feedback and modulate learning based on state-action uncertainty. For general mHealth applications, this is particularly relevant: intervention policies must be derived from limited and noisy user data, where unstable learning can lead to poor personalization or user disengagement.

Furthermore, the Bayesian ATM agent adapts its measurement strategy in a principled way. By increasing measurements in high-uncertainty states and reducing them in confident regions, the agent shows desirable behavior: focusing queries on ambiguous cases and minimizing cost or burden when unnecessary. This property is essential for long-term adherence in user deployments, where over-measuring may lead to survey fatigue or complete disengagement.

However, our results also reveal clear limitations. In highly stochastic toy environments, the Bayesian agent overcommits to measuring, incurring excessive costs without proportional benefit. This is demonstrated in the slippery variant of FrozenLake, where measurement cost dominates the reward in the scalarized return. In large-scale environments, Bayesian Q-learning suffers from sparse visitation and overgeneralization of uncertainty, leading to suboptimal exploration and performance collapse. These findings suggest that while Bayesian Q-learning improves sample efficiency in small domains, its effectiveness depends critically on accurate uncertainty calibration and sufficient data coverage.

Similarly, our results from the ADAPTS clinical testbed expose a fundamental mismatch between the original ATM assumptions and the realities of complex mHealth environments. In this setting, both ATM and its Bayesian variant fail to achieve stable learning or effective querying behavior. Unlike in the tabular domains, querying in the ADAPTS testbed has downstream effects on the environment (i.e., through engagement), the rewards are delayed and mediated by latent variables, and both the state and reward are partially observable. Additionally, the decoupled timing of control and measurement decisions—along with high-dimensional, continuous state spaces—creates a rich structure that tabular ATM methods are ill-equipped to model. These dynamics likely contributed to the persistent over-measuring and flat reward trajectories we observed.

\textbf{Future work} will focus on addressing these limitations in several ways. First, adaptive variance decay or uncertainty regularization techniques may help prevent over-measuring in low-signal regimes. Second, integrating posterior pruning or bootstrapped uncertainty methods could mitigate the overconfidence or underconfidence issues that we observed in large domains. Third, in order to improve performance in settings like ADAPTS, new algorithms must jointly model how measurements affect both state transitions and long-term rewards, requiring more accurate state and reward representations and inference mechanisms. Initial results in this research show that Bayesian Q learning methods may be promising for mHealth settings like ADAPTS to minimize user survey burden (if adapted to the ADAPTS setting, for example). Additionally, transitioning to continuous state policies and learning user-specific latent state representations is important for scaling these methods in real-world clinical applications.

Taken together, our results provide the first empirical evaluation of integrating Kalman-style Bayesian Q-learning into the Act-Then-Measure algorithm. We show that Bayesian updates substantially improve stability and learning efficiency in low-data, sparse-reward ACNO-MDP settings, precisely the configurations where replicated Q-learning struggles. However, our experiments also highlight structural limitations of ATM and its Bayesian extension when applied to more complex mHealth environments such as ADAPTS, where partial measurement, delayed burden, and stochastic engagement dynamics complicate belief updates and inflate posterior uncertainty. These findings clarify both the promise and the boundaries of Bayesian ATM, and suggest several directions for future work, including hybrid model-based/value-based updates, richer uncertainty propagation mechanisms, and algorithms tailored to the causal and behavioral characteristics of mHealth domains.

\section{Appendix}

\begin{figure}[h]
    \centering
    \includegraphics[width=0.9\linewidth]{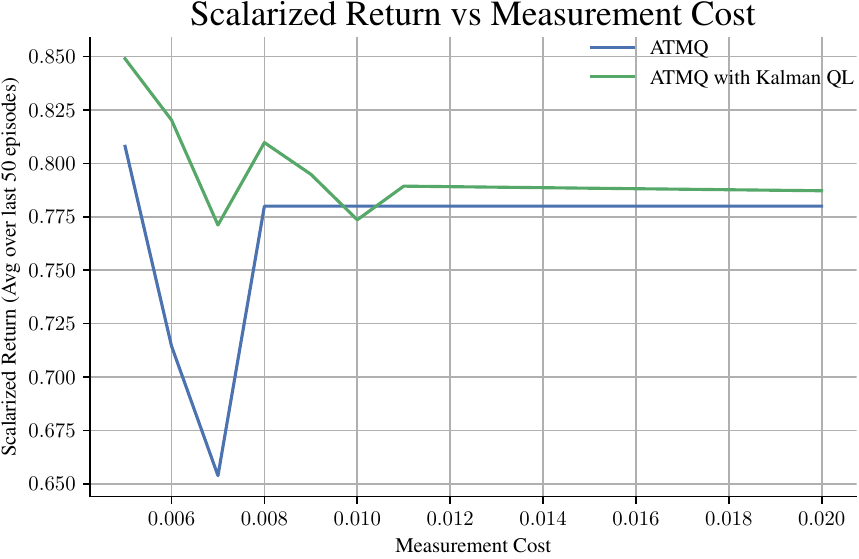}
    \caption{
        Scalarized return as a function of measurement cost for two variants of the Act-Then-Measure algorithm (ATMQ and ATMQ with Kalman Q-learning). Each point reflects the average scalarized return over the final 50 episodes, evaluated across multiple random seeds in the semi-slippery 4x4 Frozen Lake environment. As measurement cost increases, both algorithms degrade in performance, with the Bayesian QL variant exhibiting greater stability at higher costs.
    }
    \label{app:mc_plots}
\end{figure}

\bibliography{references} 
\bibliographystyle{plainnat}

\end{document}